\documentclass[conference]{IEEEtran}
\IEEEoverridecommandlockouts

\usepackage{cite}
\usepackage{comment}
\usepackage{amsmath,amssymb,amsfonts}
\usepackage{algorithmic}
\usepackage{graphicx}
\usepackage{textcomp}
\usepackage{xcolor}
\def\BibTeX{{\rm B\kern-.05em{\sc i\kern-.025em b}\kern-.08em
    T\kern-.1667em\lower.7ex\hbox{E}\kern-.125emX}}
\begin{document}

\title{A GitOps-Driven Annotation Catalog for Fully Automatic Railway Operations}

\author{
\IEEEauthorblockN{Martin Köppel}   
\IEEEauthorblockA{\textit{SIGNON Deutschland GmbH} \\
Berlin, Germany \\
martin.koeppel@deutschebahn.com}
\and
\IEEEauthorblockN{Tobias Cronauer}
\IEEEauthorblockA{\textit{DB Systel GmbH} \\
Frankfurt, Germany \\
tobias.cronauer@deutschebahn.com}
\and
\IEEEauthorblockN{Zekiye Ilknur-Öz}   
\IEEEauthorblockA{\textit{DB InfraGO AG} \\
Berlin, Germany \\
zekiye.ilknur-oez@deutschebahn.com}
\and
\IEEEauthorblockN{Sebastian Dubiel}   
\IEEEauthorblockA{\textit{DB Systel GmbH} \\
Munich, Germany \\
sebastian.dubiel@deutschebahn.com}
\and
\IEEEauthorblockN{Patrick Naumann}
\IEEEauthorblockA{\textit{SIGNON Deutschland GmbH} \\
Berlin, Germany \\
patrick.naumann@deutschebahn.com}
\and
\IEEEauthorblockN{Philipp Neumaier}
\IEEEauthorblockA{\textit{SIGNON Deutschland GmbH} \\
Berlin, Germany \\
philipp.neumaier@deutschebahn.com}
}

\maketitle

\begin{abstract}
Automatic train operation (ATO) at grade of automation 3 and above (GoA3-GoA4) requires robust AI-based perception systems capable of reliably detecting obstacles and railway-specific objects under real-world conditions. The effectiveness of these modern artificial intelligence approaches depends heavily on large-scale, high-quality, and highly dynamic annotated datasets. However, managing metadata, maintaining provenance, and tracking the iterative evolution of these annotations impose significant infrastructural and regulatory requirements. Existing monolithic data catalogs often suffer from massive operational overhead, poor integration into developer workflows, and severe documentation drift. This paper introduces an innovative, lightweight GitOps-based architecture for metadata management. By leveraging Data-as-Code principles, Continuous Integration/Continuous Deployment (CI/CD) pipelines, and Static Site Generation (SSG), the proposed approach establishes a seamless, developer-centric workflow. This ensures an traceability, enforces strict regulatory compliance, and automatically generates a highly performant dataset overview.

\end{abstract}


\section{Introduction}
\begin{figure*}[htbp]
    \centering
    \includegraphics[width=0.9\textwidth]{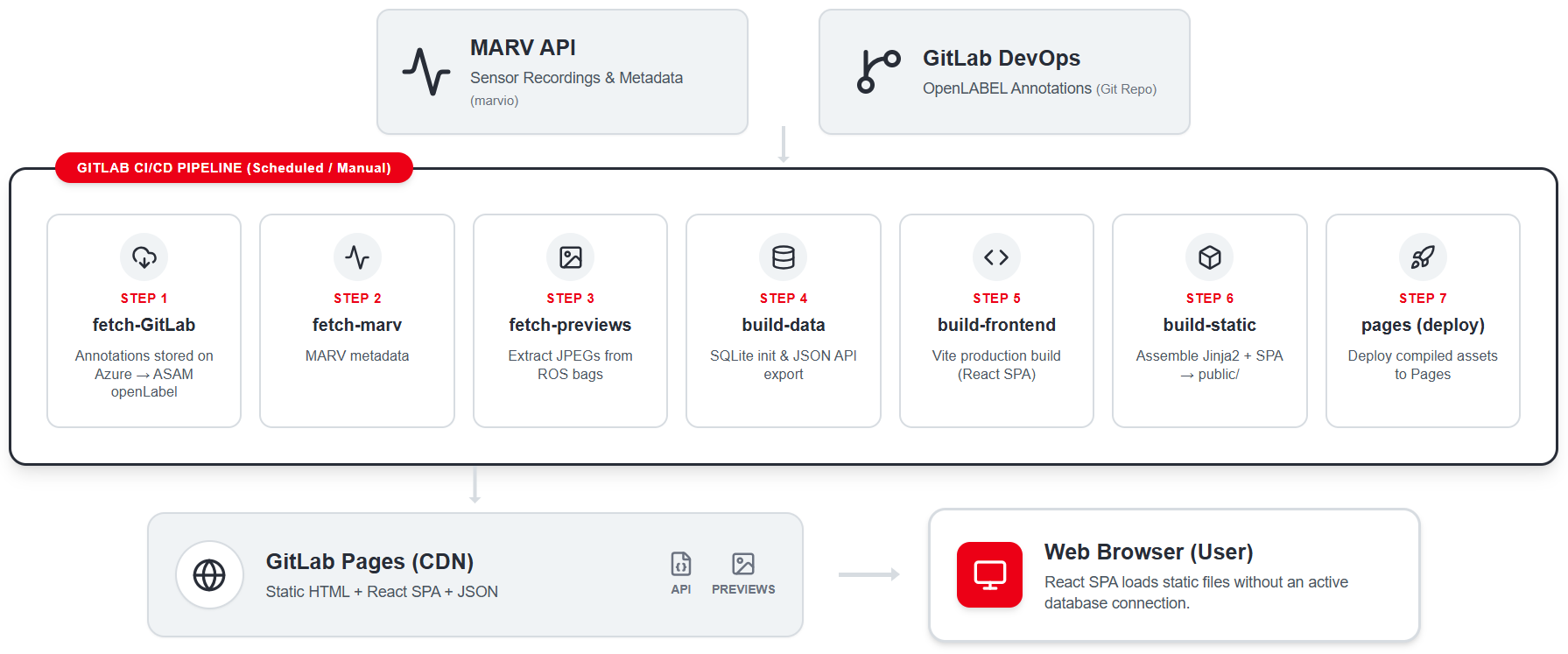}
    \caption{Overview of the annotation catalog. A serverless GitOps architecture.}
    \label{fig:overview}
\end{figure*}

The progressive deployment of automatic train operation (ATO) systems requires technical components to replace human operators. These components must reliably handle complex driving and monitoring tasks. Consequently, future autonomous and fully automatic systems depend heavily on multi-modal perception. These perception systems continuously monitor the track environment, detect obstacles, and process varying operational scenarios. Processing this multi-sensor data strictly requires artificial intelligence (AI), particularly deep neural networks. This input data typically consists of   LiDAR point clouds, high-resolution RGB and infrared cameras, radar images and GNSS data  \cite{2025-multisensor-dataset}.

Utilized AI models must achieve safety levels that equal or exceed human drivers. Therefore, they require systematic training and rigorous validation using massive amounts of domain-specific sensor data. Recent projects demonstrated the immense scale of raw data needed. For example, the multi-sensor dataset for rail vehicles proposed in \cite{2025-multisensor-dataset} provides over seven million annotations across 21 object classes. This effort builds upon earlier specialized datasets. For instance, the GERALD dataset focuses on railway signal detection \cite{2_leibner2023} and the OSDaR23 \cite{1_tagiew2023osdar23} on objects in multi-sensor data. Additionally, synthetic datasets like SynDRA address the need for domain adaptation \cite{3_damico2025}.

However, raw multi-sensory data alone has a limited usage for machine learning (ML) and neural networks. Its algorithmic value depends highly on precise annotations. Such annotations may include 2D/3D bounding boxes, 2D/3D polylines, polygons, segmentation masks, and temporally coherent tracking IDs. Currently, the efficient management, validation, and versioning of these annotations present a severe infrastructural challenge in industrial practice.

This paper introduces an annotation datacatalog. This system serves as a decentralized information hub for multi-sensor datasets. It bridges the gap between a core data management platform and GitLab DevOps. GitLab DevOps hosts the version-controlled annotations files stored in the railLabel \cite{1_tagiew2023osdar23} format. The railLabel format is a fully compatible subset of the ASAM OpenLABEL standard \cite{asam_openlabel}. The system integrates an automated quality check into a multi-stage CI/CD pipeline. This strict validation ensures that only synchronized and verified metadata is published. Furthermore, the architecture employs Static Site Generation (SSG). This approach creates a high-performance, serverless web interface. Ultimately, it allows researchers to efficiently discover, filter, and verify datasets.

\section{State-of-the-Art}

Historically, ML projects used monolithic data catalogs to manage metadata and enforce FAIR principles (Findability, Accessibility, Interoperability, Reusability).
Current literature identifies over 75 general-purpose data catalogs \cite{5_kropshofer2025}. Established systems like Amundsen (Lyft) \cite{amundsen_io} and DataHub (LinkedIn) \cite{7_garcia2025} excel at aggregating relational database structures but require significant infrastructure overhead. Amundsen relies on complex microservices and graph databases like Neo4j, while DataHub requires streaming technologies like Apache Kafka for real-time ingestion. Even established MLOps frameworks like MLflow \cite{mlflow_org}  primarily focus on model parameters rather than the high-frequency versioning of raw semantic annotations. For agile ML teams managing continuous updates of JSON annotation files, these architectures are fundamentally oversized.
Furthermore, these systems enforce a separation between the code repository and the metadata catalog. This structural divide is the primary cause of "documentation drift"—a phenomenon where manually maintained documentation increasingly diverges from the fast-evolving reality of the data pipelines \cite{8_gebru2021}, \cite{Tanetal2023}.

External data catalogs often exhibit significant rigidity. To counter this, the industry is shifting towards a "Data-as-Code" paradigm. Tools like data version control exemplify this approach. They treat datasets and schemas exactly like application code. Consequently, they establish Git as the single immutable source of truth \cite{Guptaetal2022}. These pure Data-as-Code frameworks successfully ensure cryptographic reproducibility. However, they inherently lack human-readable interfaces. Such visual accessibility remains strictly necessary for data scientists and compliance auditors.

Recent research demonstrates that SSGs can bridge the gap between static content delivery and dynamic web requirements \cite{assaf2022high}, \cite{Savenko2025}. By compiling data definitions (JSON, YAML, Markdown) into static HTML during a CI/CD process, SSGs offer a better performance and security without runtime dependencies.

Our proposed catalog interacts with the MARV \cite{marv_robotics} data management system.
MARV is a specialized data management platform designed for autonomous and fully automatic systems. It efficiently stores massive, multi-sensory recordings like ROS bags and extracts their raw metadata. However, MARV functions primarily as a backend and lacks native integration with version-controlled semantic annotations. Therefore, we implement a Git-based catalog to act as a decoupled, serverless information layer. This architecture synchronizes the raw sensor data with semantic labels, eliminates documentation drift, and enables low-bandwidth visual exploration.

\section{Overview}
The overview of the annotation datacatalog is presented in Fig.~\ref{fig:overview}.
To overcome the limitations of monolithic catalogs, we propose a lightweight, serverless data catalog tailored specifically for railway annotation datasets. The system aggregates metadata from sensor recordings (MARV API) and data annotations (GitLab DevOps) into a searchable web interface. 

Instead of operating a continuous backend server, the architecture relies entirely on GitOps principles and scheduled CI/CD pipelines. The pipeline fetches metadata, compiles a temporary SQLite database, exports it into static JSON files, and generates a React-based Single Page Application (SPA). This SPA is subsequently hosted on a Content Delivery Network (GitLab Pages).

\section{Detailed Description}

Traditional web applications typically rely on a live Relational Database Management System (RDBMS) to serve data. This conventional approach requires continuous server maintenance and consumes significant computational resources at runtime. Our technical implementation completely discards this live backend. Instead, we employ a pre-compiled static architecture. All data relationships are processed and resolved in advance during a build phase. Consequently, the system eliminates the need for an active database server when the user accesses the catalog.

\subsection{Automated Data Ingestion Pipeline}
The system relies on an automated data ingestion process. This core mechanism synchronizes two distinct data domains:

\begin{enumerate}
    \item Sensor Metadata: The system retrieves this data via the MARV API. It includes recording parameters, ROS bag details, and information about user-defined Non-Regular Scenes (NRS).
    \item Semantic Annotations: The system fetches annotations from a dedicated GitLab DevOps Git repository. They comprise JSON files that describe the annotation geometries for object classes and associated attributes. All generated JSON files comply with the railLabel specification \cite{1_tagiew2023osdar23}.
\end{enumerate}

\subsection{The 7-Stage CI/CD Publishing Process}
A strict 7-stage CI/CD pipeline orchestrates the catalog compilation. This automated process strictly guaranties data integrity:

\begin{enumerate}
    \item Linting \& Testing: The pipeline performs code quality checks using Ruff and ESLint. Afterwards, it runs over 540 automated unit tests.
    \item Fetching: The system incrementally retrieves MARV datasets and GitLab annotations.
    \item Preview Extraction: The pipeline automatically extracts preview JPEG images directly from massive ROS bags. It achieves this highly efficient extraction by utilizing HTTP range requests.
    \item Data Build: The system initializes a volatile internal SQLite database. It uses this temporary schema to join the heterogeneous sensor and annotation data. Finally, it exports this unified relational state into static JSON files.
    \item Frontend Build: The Vite build tool compiles the React and TypeScript Single Page Application (SPA).
    \item Static Build: The pipeline merges all components and assembles the final directory structure.
    \item Deployment: The system pushes the compiled static assets to GitLab Pages for hosting.
\end{enumerate}

\subsection{Technology Stack and User Interface}
The frontend of the data catalog utilizes React 18, TypeScript, and Tailwind CSS. It operates entirely as a static SPA within the browser of the user. The application uses the pre-built JSON files as its unique data source. Furthermore, the interface features a ``Sequences Page'', cf. Fig. \ref{fig:datacatalog}. This dedicated page allows researchers to efficiently filter datasets by annotation classes, specific sensors, and further tags.

\section{Results and Discussion}
\begin{figure*}[htbp]
    \centering
    \includegraphics[width=0.7\textwidth]{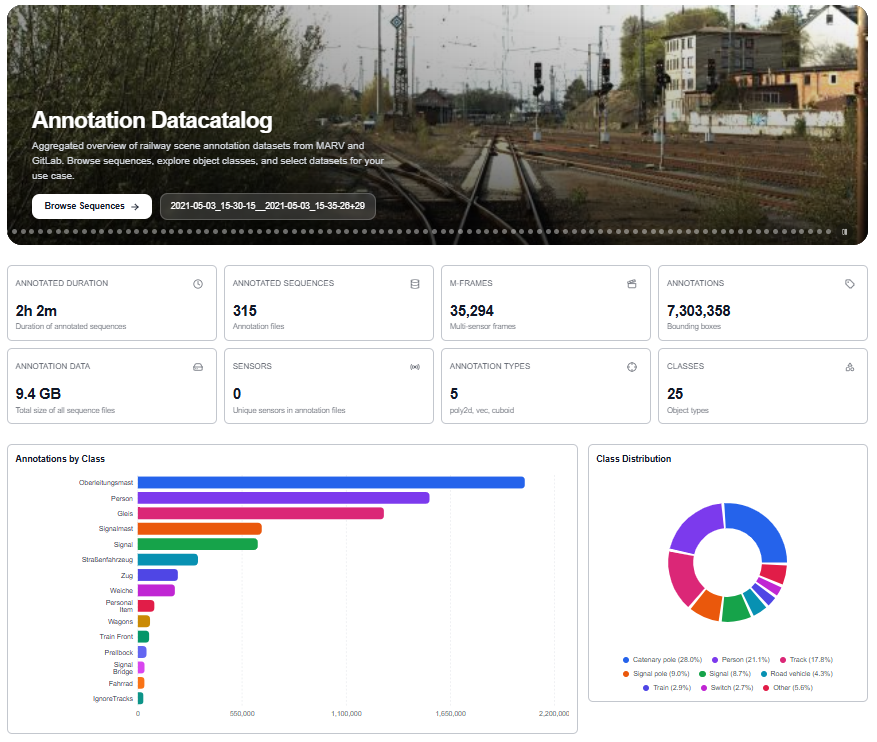} 
    \hspace*{0pt}\includegraphics[width=0.695\textwidth]{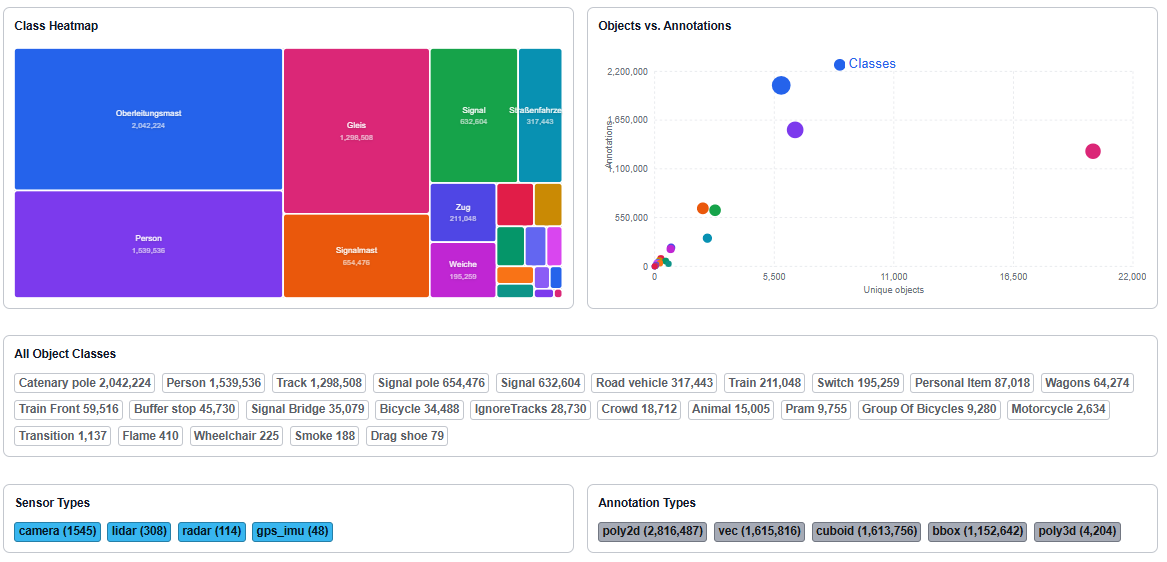} 
    \caption{The summary page of the annotation data catalog.}
    \label{fig:datacatalog}
\end{figure*}

This work shows the successful decoupling of dynamic annotations from monolithic RDBMS architectures. Using Data-as-Code, GitOps, and SSG, the proposed serverless ecosystem ensures regulatory compliance and development agility. The resulting interface of the proposed annotation data catalog is shown in Fig. \ref{fig:datacatalog}.

Performance evaluations of the proposed architecture demonstrate page load times between 138 and 414 ms and CI/CD build durations between 20:07 and 49:13 minutes. These metrics serve as baseline indicators rather than absolute benchmarks. Actual build times depend heavily on shared GitLab runner resources, while perceived load times vary significantly based on client hardware, network conditions, and caching mechanisms.

In the following the proposed architecture is evaluated against current state-of-the-art solutions. Comparing the GitOps-driven catalog with monolithic systems reveals distinct advantages. These benefits are very important for fully automatic railway operations. 

\subsection{Comparative Evaluation against Established Monoliths}

Empirical studies indicate that the adoption rate of monolithic enterprise data catalogs frequently remains below 30\% \cite{Singhetal2025}. This underutilization primarily stems from a fundamental friction between rigid catalog infrastructures and agile developer workflows. To contextualize the advantages of our proposed GitOps architecture, Table \ref{tab:arch_comparison} systematically compares it against these traditional systems.

\begin{table*}[htbp]
    \centering
    \caption{Architectural Comparison of Metadata Management Systems}
    \label{tab:arch_comparison}
    \renewcommand{\arraystretch}{1.3} 
    \begin{tabular}{|p{3cm}|p{2.5cm}|p{2.5cm}|p{2.5cm}|p{3.5cm}|}
        \hline
        \textbf{Evaluation Criteria} & \textbf{Amundsen} & \textbf{DataHub} & \textbf{MLflow} & \textbf{GitOps + SSG Catalog (Proposed)} \\
        \hline
        Architecture Type & Modular, Graph-based & Push-based, Streaming & Tracking-Server, RDBMS & Serverless, Static Site Generation \\
        \hline
        Infrastructure Overhead & High (Neo4j, Elasticsearch) & Very High (Kafka, DBs) & Moderate (PostgreSQL) & Minimal (Uses Git \& CI/CD) \\
        \hline
        Documentation Drift & High (Manual sync required) & Moderate to High & High (Logging chaos risk) & Zero (Generated deterministically) \\
        \hline
        Primary Focus & Enterprise Data Discovery & Enterprise Lineage & Experiment Tracking & High-Frequency Raw Data \& Annotations \\
        \hline
    \end{tabular}
\end{table*}

\subsection{Amundsen: Infrastructure Over-provisioning vs. Serverless Agility}

Amundsen \cite{amundsen_io} employs a complex microservices architecture that necessitates dedicated backend services, an Elasticsearch index for full-text search, and Neo4j for relationship mapping \cite{Elouataoui2026}. The requirement to operate parallel stateful databases introduces significant infrastructure costs and maintenance complexity. For agile teams primarily focused on the high-frequency versioning of JSON annotations, this design represents a disproportionate operational overhead. Furthermore, reports of a dormant development roadmap render the platform increasingly inviable for modern, fast-paced AI environments \cite{Elouataoui2026}.

Our proposed system eliminates this runtime complexity. Instead of maintaining Neo4j and Elasticsearch, relational logic is shifted to the build phase. The CI/CD pipeline utilizes a temporary SQLite database to aggregate data relations before discarding it. Deploying the catalog as a static React SPA via GitLab Pages reduces server maintenance costs to zero. Furthermore, this architecture physically eliminates the attack surface for common cyber threats, such as SQL injections.

\subsection{DataHub: Cognitive Overload vs. Domain-Specific Focus}

DataHub employs a push-based architecture designed for petabyte-scale streaming via Apache Kafka. This architecture can lead to excessive resource consumption during setup. Furthermore, extensive lineage graphs often create cluttered visualizations \cite{7_garcia2025}. These complex views significantly increase the cognitive load for data scientists.

In contrast, the proposed annotation data catalog prioritizes functional utility and domain-specific requirements. The system avoids mapping 
corporate data flows. Instead, it focuses exclusively on assets critical for automatic train operation. These include raw sensor recordings and structured annotations. Rather than requiring complex push APIs, the catalog utilizes a standard Git workflow. The documentation is created automatically as an inherent result of the version control commit cycle. This approach successfully resolves the conflict between developmental agility and documentation discipline.

\subsection{MLflow: Logging Chaos vs. Strict Schema Enforcement}

MLflow \cite{mlflow_org} is a leading tool for experiment tracking, yet it exhibits limitations when versioning massive raw datasets at high frequencies. The absence of strict schema enforcement frequently results in ``logging chaos'' \cite{Micallef2026}. This lack of standardization creates incompatible experiments due to inconsistent naming conventions, such as using both ``accuracy'' and ``acc.'' Furthermore, MLflow's user interface suffers from performance degradation during dense experimental runs and requires continuous maintenance of a stateful PostgreSQL backend.

In contrast, our GitOps-driven catalog ensures robust metadata management by enforcing type safety and schema conformity before data is ingestioned. A comprehensive CI/CD pipeline serves as a validation gate, employing static analysis and automated tests to ensure the integrity of all ingested metadata. While MLflow focuses on the final trained model at the end of the pipeline, our architecture ensures the integrity of the input material at the source.

\subsection{Eliminating ``Documentation Drift'' and Ensuring Compliance}

Utilizing the Git commit of the annotation file as the trigger for the CI/CD documentation pipeline completely eliminates documentation drift \cite{Tanetal2023}.
The pipeline serves as a strict validation gate. Any missing or invalid metadata results in a build failure, thereby preventing non-compliant data from entering the catalog. For compliance auditing under the European AI Act \cite{6_euai2024}, the cryptographically secure Git history provides an immutable audit trail of every data modification.

\subsection{Resource-Optimized Data Exploration}

The management of autonomous and fully automatic driving data presents a significant challenge due to the massive volume of Robot Operating System (ROS) bag files. The proposed architecture optimizes data exploration by utilizing HTTP Range Requests to extract specific JPEG preview images directly from binary streams. This mechanism enables the retrieval of visual metadata without necessitating the download of multi-gigabyte recordings. Consequently, the data transfer requirements are reduced to a few megabytes per dataset. This approach significantly decreases network bandwidth consumption and improves the temporal efficiency of pipeline execution.

\section{Conclusion and Future Work}

In this work a GitOps-driven annotation data catalog is presented.
The effective decoupling of dynamic machine learning (ML)  annotations from monolithic relational database (RDBMS) architectures is demonstrated. We introduce a robust, serverless ecosystem for dataset management. The system integrates Data-as-Code principles, stringent GitOps pipelines, and Static Site Generation (SSG). This framework addresses the compliance and auditability standards for contemporary artificial intelligence (AI) regulations. Furthermore, the architecture preserves necessary development agility for data science teams.
Future research will focus on expanding the automated CI/CD validation mechanisms. 

\section*{Acknowledgment}
The authors would like to thank Martin Boekhoff, Volker Eiselein, Erik Bochinski (SIGNON Deutschland GmbH) and Florian Friesdorf (Ternaris)  for their valuable contributions.

\IEEEtriggeratref{1}
\IEEEtriggercmd{\vspace{+\baselineskip}}
\bibliographystyle{IEEEtran}
\bibliography{IEEEexample}

\end{document}